\documentclass[10pt,twocolumn,letterpaper]{article}
\pdfoutput=1

\usepackage[pagenumbers]{cvpr} 
\usepackage{algorithm}
\usepackage{algorithmic}
\usepackage{amsmath, amsfonts, amssymb}
\usepackage{multirow} 
\usepackage{graphicx}
\usepackage{listings}

\definecolor{cvprblue}{rgb}{0.21,0.49,0.74}
\usepackage[pagebackref,breaklinks,colorlinks,allcolors=cvprblue]{hyperref}

\def\paperID{3296} 
\def\confName{CVPR}
\def\confYear{2025}

\title{Gaussians on their Way: Wasserstein-Constrained 4D Gaussian Splatting with State-Space Modeling}

\author{Junli Deng\\
Communication University of China\\
Beijing, China\\
{\tt\small dengjunliok@cuc.edu.cn}
\and
Yihao Luo\\
Imperial College London\\
London, UK\\
{\tt\small y.luo23@imperial.ac.uk}
}

\begin{document}
\maketitle
\begin{abstract}
Dynamic scene rendering has taken a leap forward with the rise of 4D Gaussian Splatting, but there’s still one elusive challenge: how to make 3D Gaussians move through time as naturally as they would in the real world, all while keeping the motion smooth and consistent. In this paper, we unveil a fresh approach that blends state-space modeling with Wasserstein geometry, paving the way for a more fluid and coherent representation of dynamic scenes. We introduce a State Consistency Filter that merges prior predictions with the current observations, enabling Gaussians to stay true to their way over time. We also employ Wasserstein distance regularization to ensure smooth, consistent updates of Gaussian parameters, reducing motion artifacts. Lastly, we leverage Wasserstein geometry to capture both translational motion and shape deformations, creating a more physically plausible model for dynamic scenes. Our approach guides Gaussians along their natural way in the Wasserstein space, achieving smoother, more realistic motion and stronger temporal coherence. Experimental results show significant improvements in rendering quality and efficiency, outperforming current state-of-the-art techniques.
\end{abstract}
    

\section{Introduction}

Dynamic scene rendering is a fundamental problem in computer vision, with widespread applications in virtual reality, augmented reality, robotics, and film production. Accurately capturing and rendering dynamic scenes with complex motions and deformations remains a challenging task due to the high computational demands and the intricate nature of dynamic environments~\cite{wang2024deep,HUANGJiahui_14}.

Neural representations have advanced dynamic scene modeling, with Neural Radiance Fields~\cite{mildenhall2020nerf} revolutionizing novel view synthesis through neural network-parameterized continuous functions. Extensions to dynamic scenes~\cite{park2021nerfies, li2021neural,Yan_2023_CVPR,park2021hypernerf,shao2023tensor4d,Point-DynRF} have been proposed, but they often suffer from high computational costs and limited real-time capabilities. 4D Gaussian Splatting~\cite{wu20244d,yang2024deformable,duan20244d,yangreal,li2024spacetime,luiten2023dynamic} enables real-time dynamic scene rendering using dynamic 3D Gaussians and differentiable splatting~\cite{kerbl3Dgaussians}. However, accurately modeling scene dynamics remains challenging due to limitations in estimating precise Gaussian transformations~\cite{wu2024recent,fei20243d}.

In this paper, we draw inspiration from control theory~\cite{catlin2012estimation} and propose a novel approach that integrates a State Consistency Filter into the 4D Gaussian Splatting framework. By modeling the deformation of each Gaussian as a state in a dynamic system, we estimate Gaussian transformations by merging prior predictions and observed data, accounting for uncertainties in both.

To ensure smooth and consistent parameter updates, we incorporate Wasserstein distance~\cite{givens1984class,panaretos2019statistical} as a key metric between Gaussian distributions. This metric effectively quantifies the optimal transformation cost between distributions, considering both positional and shape differences. By using Wasserstein distance regularization, we preserve the underlying Gaussian structure while enhancing temporal consistency and reducing rendering artifacts.

Additionally, we introduce Wasserstein geometry~\cite{ambrosio2008gradient,panaretos2019statistical} to model Gaussian dynamics, capturing both translational motion and shape deformations in a unified framework. This approach enables more physically plausible evolution of Gaussians, leading to improved motion trajectories and rendering quality. Our main contributions are:
\begin{itemize}
    \item We propose a novel framework that integrates a State Consistency Filter into 4D Gaussian Splatting, enabling more accurate Gaussian motion estimation by optimally merging prior predictions and observed data.
    \item We introduce Wasserstein distance regularization, which smooths Gaussian parameter updates over time, ensuring temporal consistency and reducing artifacts.
    \item We leverage Wasserstein geometry to model both translational motion and shape deformations of Gaussians, enhancing the physical plausibility of Gaussian dynamics.
\end{itemize}

\begin{figure*}[!htbp]
    \centering
    \includegraphics[width=0.9\textwidth]{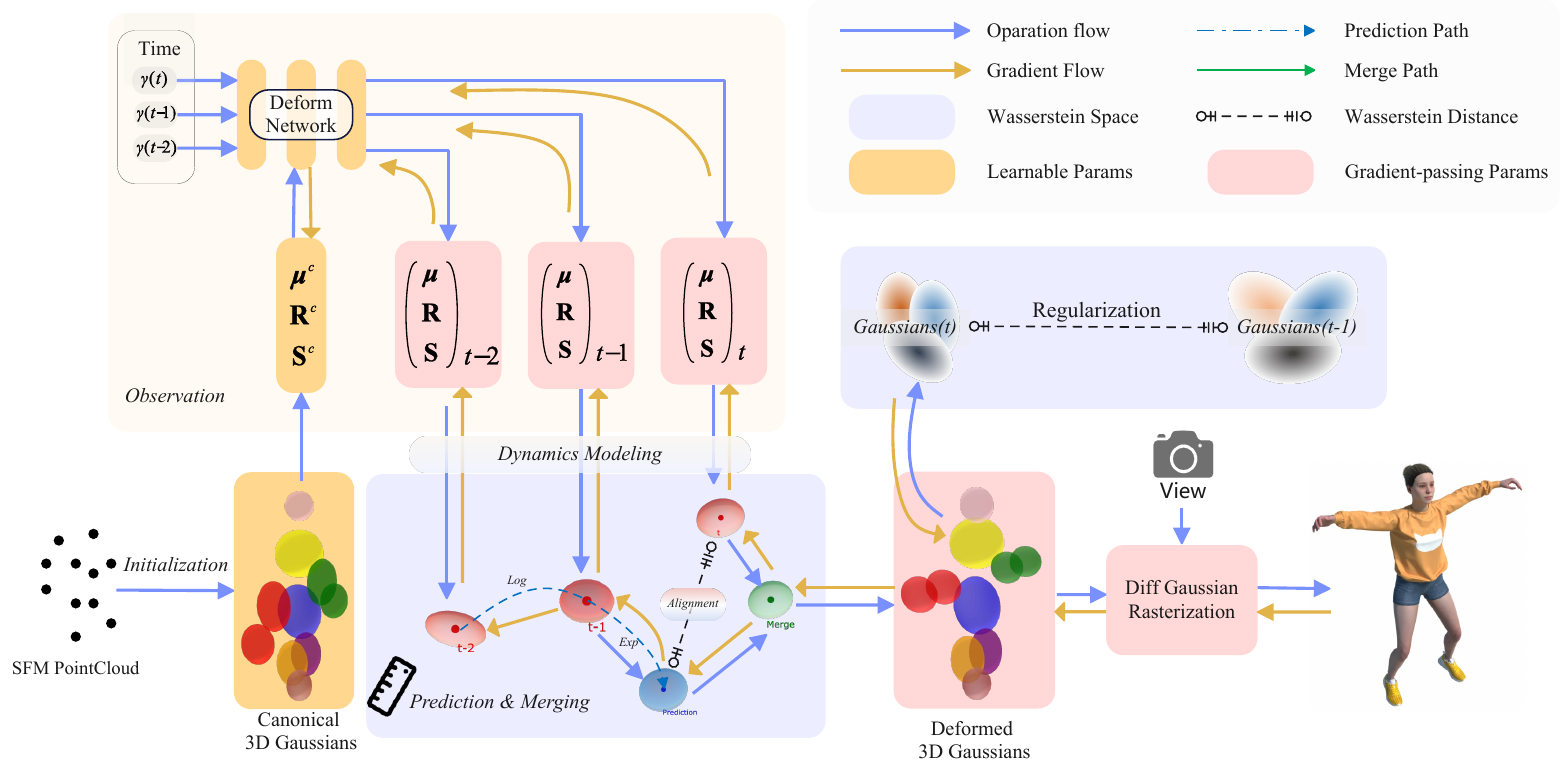} 
    \caption{Overview of our proposed method. Starting from a Structure-from-Motion (SFM) point cloud, we initialize canonical 3D Gaussians including position $\boldsymbol{\mu}^c$, rotation $\mathbf{R}^c$, and scale $\mathbf{S}^c$ parameters. The deform network predicts these parameters $(\boldsymbol{\mu},\mathbf{R},\mathbf{S})$ at different timestamps $\gamma(t)$. In the Wasserstein space, our state-updating mechanism merges predictions with observations, while ensuring temporal coherence between frames by regularization. The merged Gaussians are then rendered via differentiable rasterization.}
    \label{fig:method_overview}
\end{figure*}


\section{Related Work}

\subsection{Dynamic Novel View Synthesis}

Synthesizing new views of dynamic scenes from multi-time 2D images remains challenging. Recent works have extended Neural Radiance Fields (NeRF) to handle dynamic scenes by learning spatio-temporal mappings~\cite{Lombardi2019, mildenhall2019llff, pumarola2021d, li2022neural, li2022streaming, cao2023hexplane, song2023nerfplayer, attal2023hyperreel, fridovich2023k, Wang2023ICCV}. While classical approaches using the plenoptic function~\cite{adelson1991plenoptic}, image-based rendering~\cite{levoy1996light, buehler2001unstructured}, or explicit geometry~\cite{riegler2020free, riegler2021stable} face memory limitations, implicit representations~\cite{park2021nerfies, wu2022d, weng2022humannerf, li2022neural, gao2021dynamic} have shown promise through deformation fields~\cite{pumarola2021d, park2021nerfies, park2021hypernerf} and specialized priors~\cite{weng2022humannerf, alldieck2021imghum, jiang2023instantavatar, athar2022rignerf, bai2023high, peng2021animatable}.

Temporally extended 3D Gaussian Splatting has also been explored for dynamic view synthesis. Luiten~\etal~\cite{luiten2023dynamic} assign parameters to 3D Gaussians at each timestamp and use regularization to enforce rigidity. Yang~\etal~\cite{yang2023real} model density changes over time using Gaussian probability to represent dynamic scenes. However, they require many primitives to capture complex temporal changes. Other works~\cite{wu20234dgaussians, yang2023deformable, huang2023sc, qian20233dgsavatar, hu2023gauhuman, lu20243d} leverage Multi-Layer Perceptrons (MLPs) to represent temporal changes. In 4D Gaussian Splatting, the motion of Gaussians should adhere to physical laws. By incorporating control theory, we can predict the motion of Gaussians more accurately.

\subsection{Dynamic Scene State Estimation}

Recent advances in dynamic scene reconstruction have explored various methods for tracking and modeling temporal changes. Traditional approaches, such as SORT~\cite{bewley2016simple} and SLAM systems~\cite{davison2007monoslam}, provide robust state estimation frameworks. Learning-based techniques further enhance these methods, particularly in handling complex scenarios with limited observations~\cite{wagstaff2022self, revach2022kalmannet}.

Recent studies have investigated probabilistic and optimal transport-based approaches for dynamic scenes. GaussianCube~\cite{zhang2024gaussiancube} models scenes using probabilistic distributions to handle deformations robustly, while Shape of Motion~\cite{wang2024shapemotion4dreconstruction} leverages geometric transformations for temporal coherence. KFD-NeRF~\cite{zhan2023kfd} applies Kalman filtering to NeRF but is limited by its discrete point representation, which lacks the geometric structure of optimal transport. In contrast, our method employs 3D Gaussian splatting in Wasserstein space, integrating both position and covariance (i.e., shape) as fundamental components of each Gaussian distribution. This enables smooth updates by evolving distributions across frames on a Wasserstein manifold. Optimal transport has also shown promise in improving dynamic NeRF convergence in OTDNeRF~\cite{ramasinghe2024improving}, which applies unconstrained Wasserstein transport to rendered images or latent spaces, our approach directly considers the Wasserstein distance between dynamic Gaussians.

Our method distinguishes itself by leveraging Gaussian splatting to model dynamic elements as full probability distributions in Wasserstein space. This formulation captures the geometric nature of distribution transformations, making it particularly effective for scenes with significant deformations or rapid motion.


\section{Method}
Our framework integrates three key components for dynamic scene rendering (Figure~\ref{fig:method_overview}). Sec.~\hyperref[sec:filter]{3.1} provides a simple Euclidean-based framework for building the state space for Gaussians across frames, assuming linear motion for Gaussians. However, handling covariance as separate elements is flawed; covariance must be treated as a unified entity. Thus, in Sec.~\hyperref[sec:regularization]{3.2}, we introduce Wasserstein distance as a proper distributional metric and the Wasserstein distance algorithm for regularization, and in Sec.~\hyperref[sec:velocity_computation]{3.3}, we extend it to the Wasserstein geometry on the Gaussian manifold, enabling smooth, physically intuitive Gaussian dynamics.

\subsection{Filter for State Consistency}
\label{sec:filter}
\subsubsection{Observer: Neural Gaussian Deformation Field}

3D Gaussian Splatting represents static scenes as a collection of 3D Gaussians, each parameterized by its mean position $\boldsymbol{\mu}$ and covariance matrix $\boldsymbol{\Sigma}$. The covariance matrix is typically decomposed into rotation $R$ and scaling $S$ matrices~\cite{kerbl3Dgaussians}:
\begin{equation}
    \boldsymbol{\Sigma} = \mathbf{R}  \mathbf{S} \mathbf{S}^T \mathbf{R}^T.
\label{RSSR}
\end{equation}

This decomposition allows for efficient modeling of oriented Gaussian distributions in 3D space. 4D Gaussian Splatting extends this representation to dynamic scenes by allowing these Gaussian parameters to vary over time $\gamma(t)$, enabling the modeling of moving and deforming objects.

Building upon this foundation, we introduce a more principled approach to modeling temporal variations through a neural deformation field. Given a canonical Gaussian distribution $\mathcal{N}^c = \mathcal{N}(\boldsymbol{\mu}^c, \boldsymbol{\Sigma}^c)$ and a time parameter $t$, our neural deformation field predicts the observed Gaussian distribution $\mathcal{N}_t^{Ob}(\boldsymbol{\mu}_t^{Ob}, \boldsymbol{\Sigma}_t^{Ob})$ at time $t$:

\begin{equation}
    \mathcal{N}_t^{Ob} = f_{\theta}(\mathcal{N}^c, t),
\end{equation}
where $\theta$ represents the learnable parameters of the neural network $f_{\theta}$ implemented as a Multi-Layer Perceptron (MLP). $f_{\theta}$ takes the concatenation of the canonical Gaussian parameters and the positional time encoding~\cite{mildenhall2020nerf} as input and outputs the transformation parameters that map the canonical Gaussian to the observed state:
\begin{equation}
\begin{aligned}
    \boldsymbol{\mu}_t^{Ob} &= \boldsymbol{\mu}^c + \Delta\boldsymbol{\mu}_t, \\
    \boldsymbol{\Sigma}_t^{Ob} &= \boldsymbol{\Sigma}^c + \Delta\boldsymbol{\Sigma}_t,
\end{aligned}
\end{equation}
where $\Delta\boldsymbol{\mu}_t$ is the translation offset and $\Delta\boldsymbol{\Sigma}_t$ is the deformation of the covariance matrix.

Using deformed Gaussian distributions $\mathcal{N}_t^{Ob}$ predicted by the neural deformation field to represent dynamic scenes is a common practice in 4D Gaussian Splatting frameworks~\cite{wu20244d,yang2024deformable,duan20244d,yangreal,li2024spacetime,luiten2023dynamic}. However, these methods often suffer from flickering artifacts due to abrupt changes in Gaussian parameters between frames. To address this issue, we use the above deformed Gaussian distributions as observations in a state consistency filter, which merges the predicted states with the observed data to obtain the final Gaussian parameters for rendering.

\subsubsection{Predictor: Time-Independent Linear Dynamics}

Traditional Kalman Filters~\cite{kalman1960new} model the state evolution as a linear dynamical system, where the state at time $t$ is a linear transformation of the state at time $t-1$ combined the control input at time $t$. The distribution of the state is updated based on the observed data and the predicted state. 
In our case, we directly model the Gaussian distributions as states (no distribution of states is considered) with the mean and covariance as the state variables. The Euclidean state transition is given by
\begin{equation}
\begin{aligned}
    &\mathcal{N}^P_{t+1} = \mathcal{N}_t + \mathbf{v}_t \Delta t, \\
    &\mathbf{v}^P_t = \mathcal{N}_t - \mathcal{N}_{t-1},\\
    & \mathbf{v}^P_t = \mathbf{v}_{t-1},
\end{aligned}
\label{prediction gaussian}
\end{equation}
where $\mathcal{N}^P_{t+1}$ is the predicted Gaussian distribution at time $t+1$, $\mathcal{N}_t$ is the Gaussian distribution at time $t$, $\mathbf{v}_t$ is the velocity of the Gaussian at time $t$, 
and $\Delta t =1$ is the time step. In Euclidean metric, the velocity $\mathbf{v}_t$ can be decomposed into Euclidean difference of means and covariances, i.e.,
$$
\mathcal{N}_{t+1} - \mathcal{N}_t = \mathbf{v}_t = (\boldsymbol{\mu}_t - \boldsymbol{\mu}_{t-1}, \boldsymbol{\Sigma}_t - \boldsymbol{\Sigma}_{t-1}).
$$

Conventionally, The predicted velocity $\mathbf{v}^P_t$ is computed as the Euclidean difference between the Gaussian distributions at time $t$ and $t-1$. Similarly, the first equation in \eqref{prediction gaussian} only considers the first-order linear dynamic in Euclidean space. In Section~\ref{sec:velocity_computation}, we will introduce the Wasserstein dynamic of Gaussian distributions to replace the Euclidean one for a better depiction of 4D Gaussian splitting. Abstractly, Wasserstein difference is defined as 
\begin{equation}
\begin{aligned}
    &\mathbf{v}_t = -\log_{\mathcal{N}_t}(\mathcal{N}_{t-1}), \\
    & \mathcal{N}_{t+1} = \exp_{\mathcal{N}_t}(\mathbf{v}_t),
\end{aligned}
\end{equation}
where the Exponential $exp$ maps a tangent (velocity) vector to an endpoint Gaussian, and Logarithm $log$ does the inverse side, assigning the endpoint Gaussian to a tangent vector. Exponential and Logarithm will be determined by the Riemannian metric endowed on the manifold of all Gaussian distributions.

Notice that both the predicted Gaussian distribution and velocity contain components for position and covariance. In the above model, we assume that the acceleration of the Gaussian distribution vanishes and the velocity remains constant over time for smoothness. The dynamics of the Gaussian distributions are modeled as a naive linear system, which is an oversimplified model and far from the real-world dynamics, but provides higher robustness for 4D Gaussian Splatting. Subsequently, we introduce a Kalman-like state updating mechanism to refine the Gaussian distributions based on the observed data and the predicted state.

\subsubsection{Merging: Kalman-like State
Updating}

The Kalman Filter~\cite{kalman1960new} is a recursive algorithm that estimates the state of a linear dynamical system from a series of noisy observations. It combines prior predictions with new measurements to produce optimal state estimates, accounting for uncertainties in both the process and the observations. Learning from Kalman-like filters, we designed a fusion mechanic to merge two Gaussians: one predicted by the Wasserstein dynamic on previous stages $(t-2,t-1)$ and the other newly observed by the network at $ t$. We do not adopt distinct process/measurement covariances for sensor noise of location vectors. Instead, we treat each Gaussian as a whole element and merely inherit ``Kalman gain" as weights to take the balance from prediction and observation. In other words, there is no noise distribution for prediction or observation but a simple merging of the two Gaussians. 

In our context, our prediction and observation are the Gaussian distributions themselves. The counterbalancing of prior predictions and new observations allows for robust tracking of the Gaussian states over time, enabling accurate rendering of dynamic scenes. We directly apply the updated equations of the Kalman Filter to merge the predicted Gaussian $\mathcal{N}_t^{Ob}$ distributions with the observed data $\mathcal{N}_t^{P}$ to obtain the updated Gaussian distributions $\hat{\mathcal{N}}_t$:

\begin{equation}
    \begin{aligned}
    & K = \boldsymbol{\Sigma}_t^{Ob} (\boldsymbol{\Sigma}_t^{Ob} + \boldsymbol{\Sigma}_t^{P})^{-1}, \\
    & \hat{\mathcal{N}}_t = \mathcal{N}_t^{Ob} + K (\mathcal{N}_t^{P} - \mathcal{N}_t^{Ob}),
    \end{aligned}
\end{equation}
where $K$ is the Kalman Gain. The Kalman Gain determines the weight given to the new observation relative to the prior prediction. A higher gain gives more weight to the observation, while a lower gain relies more on the prior prediction. The updated Gaussian distributions $\hat{\mathcal{N}}_t$ determine the final 3D representation at time $t$ and are used to render the result RGB images.

\subsection{Wasserstein Regularization}
\label{sec:regularization}
4D Gaussian Splatting essentially updates the parameters of 3D Gaussian distributions based on different input timestamps. Ensuring consistent and smooth updates of these parameters is crucial for high-quality dynamic scene rendering. We hypothesize that flickering artifacts arise when Gaussian distributions undergo abrupt changes in shape or position between consecutive frames.

Previous methods have attempted to constrain these frame-to-frame changes using simple Euclidean metrics. Some works~\cite{huang2023sc,yang2023deformable} apply Euclidean distance regularization on Gaussian means, while others either ignore covariance updates or use the Frobenius norm for regularization~\cite{bickel2008regularized}. However, these approaches treat position and shape parameters independently, failing to capture the intrinsic geometric relationship between Gaussian distributions, leading to suboptimal results. Intuitively, instead of updating the 9D parameters (3D mean and 6D covariance) in a Euclidean manner, it is more reasonable to consider the 3D Gaussian distribution as a whole and update it accordingly.

As a solution, we leverage the Wasserstein distance~\cite{panaretos2019statistical} from optimal transport theory~\cite{villani2021topics}. This metric is particularly suitable as it naturally captures both position and shape changes of Gaussian distributions by measuring the optimal mass transportation cost between them. Unlike Euclidean metrics that treat parameters independently, the Wasserstein distance provides a geometrically meaningful way to track the evolution of 3D Gaussians in dynamic scenes.

Specifically, the squared 2-Wasserstein distance between two Gaussian distributions $\mathcal{N}_1(\boldsymbol{\mu}_1, \boldsymbol{\Sigma}_1)$ and $\mathcal{N}_2(\boldsymbol{\mu}_2, \boldsymbol{\Sigma}_2)$ is given by~\cite{givens1984class}:
\begin{equation}
\label{eq:wasserstein_gaussian}
W_2^2 = \|\boldsymbol{\mu}_1 - \boldsymbol{\mu}_2\|^2 + \operatorname{Tr}(\boldsymbol{\Sigma}_1 + \boldsymbol{\Sigma}_2 - 2{(\boldsymbol{\Sigma}_1\boldsymbol{\Sigma}_2)}^{\frac{1}{2}}),
\end{equation}
where $\operatorname{Tr}({(\boldsymbol{\Sigma}_1\boldsymbol{\Sigma}_2)}^{\frac{1}{2}}) = \operatorname{Tr}({( \boldsymbol{\Sigma}_2^{\frac{1}{2}} \boldsymbol{\Sigma}_1 \boldsymbol{\Sigma}_2^{\frac{1}{2}})}^{\frac{1}{2}})$ provides a symmetric version for stable computation. The first term quantifies the squared Euclidean distance between means, and the trace term measures covariance differences in $3$-dimensional symmetric positive definite manifold $\mathrm{SPD}(3)$. This formulation captures the geometric and statistical `distance' between the distributions, providing a comprehensive measure of their disparity.

Notably, the trace term in Eq.~\eqref{eq:wasserstein_gaussian} is isometric under similarity transformations~\cite{luo2021geometric}. For 3D Gaussian Splatting with covariance matrices decomposed into rotation ($\mathbf{R}_1$, $\mathbf{R}_2$) and scale ($\mathbf{S}_1$, $\mathbf{S}_2$) matrices as Eq.~\eqref{RSSR}, the trace term becomes:
\begin{equation}
\label{eq:wasserstein_gaussian_decompose_1}
\begin{aligned}
D_{\Sigma} &= \operatorname{Tr}\left( \mathbf{S}_1 + \mathbf{S}_2 - 2\left( \mathbf{S}_1^{1/2} \mathbf{E}_{12} \mathbf{S}_1^{1/2} \right)^{1/2} \right), \\
\mathbf{E}_{12} &= \mathbf{R}_1^\intercal \mathbf{R}_2\, \mathbf{S}_2\, \mathbf{R}_2^\intercal \mathbf{R}_1,
\end{aligned}
\end{equation}
where $\mathbf{S}_1^{1/2}$ is the square root of the diagonal scale matrix $\mathbf{S}_1$, and $\mathbf{E}_{12}$ is the covariance matrix of the transformed distribution $\mathcal{N}_2$ under the rotation $\mathbf{R}_1$. This decomposition allows for computationally efficient and stable computation of the matrix square root and eigenvalue decomposition required in the Wasserstein distance calculation. The detailed implementation is provided in Algorithm~\ref{alg:wasserstein_gaussian}.

\begin{algorithm}[h!]
  \caption{Wasserstein Distance for 3D Gaussians}
  \hspace*{0.02in} {\bf Input:} Two Gaussians $\mathcal{N}_i =(\boldsymbol{\mu}_i, \mathbf{S}_i, \mathbf{R}_i), i =1,2$, \\
  \hspace*{0.02in} {\bf Output:} Wasserstein distance $W_2(\mathcal{N}_1,\mathcal{N}_2)$, \\
  \begin{algorithmic}[1]
    \STATE Euclidean difference of means $D_{\mu}^2 = \|\boldsymbol{\mu}_1 - \boldsymbol{\mu}_2\|^2$
    \STATE Compute $\operatorname{Tr}((\boldsymbol{\Sigma}_1\boldsymbol{\Sigma}_2)^{1/2})$ by 
    
    $\mathbf{E}_{12} = \mathbf{R}_1^\intercal \mathbf{R}_2 \mathbf{S}_2 \mathbf{R}_2^\intercal \mathbf{R}_1,$

    $\mathbf{C}_{12} = \mathbf{S}_1^{1/2} \mathbf{E}_{12} \mathbf{S}_1^{1/2},$

    Re-symmetrize (Optional): $\mathbf{C}_{12} \leftarrow \frac{1}{2}(\mathbf{C}_{12} + \mathbf{C}_{12}^\intercal),$

    Eigenvalue decomposition $\mathbf{C}_{12}\mathbf{e}_k = \lambda_k \mathbf{e}_k,$

    $\operatorname{Tr}((\boldsymbol{\Sigma}_1\boldsymbol{\Sigma}_2)^{1/2}) = \operatorname{Tr}(\mathbf{C}_{12}^{1/2})= \sum_k \sqrt{\lambda_k}.$ 
    \STATE Get the distance $W_2^2 = D_{\mu}^2 + \sum(\mathbf{S}_1+\mathbf{S}_2) -2 \sum_k \sqrt{\lambda_k},$
    
    Clamp (Optional) $W_2^2 \geq 0$ 

    \RETURN $W_2(\mathcal{N}_1,\mathcal{N}_2) = \sqrt{W_2^2}$
  \end{algorithmic}
  \label{alg:wasserstein_gaussian}
\end{algorithm}



We incorporate the Wasserstein distance into our optimization framework through two complementary losses. The first, our State-Observation Alignment Loss (SOA Loss), enforces physical motion consistency:
\begin{equation}
\mathcal{L}_{\text{SOA}} = W_2^2\left( \mathcal{N}^P_t,\, \mathcal{N}^{\text{Ob}}_t \right),
\label{eq:wasserstein_loss}
\end{equation}
which encourages the predicted Gaussians to align with observations while maintaining physical plausibility. While observations are inherently error-prone due to discrete temporal sampling, our predictions incorporate prior knowledge of kinematic models. By measuring the Wasserstein distance between predictions and observations, we ensure that our predicted states remain physically coherent while staying close to the observed data.

Secondly, we introduce a Wasserstein regularization term to ensure temporal consistency and mitigate artifacts between consecutive frames for all Gaussians:
\begin{equation}
\label{eq:wasserstein_regularization}
\mathcal{L}_{\text{WR}} = \sum_{t}^T\sum_{i}^N W_2^2( \hat{\mathcal{N}}^{(i)}_t, \hat{\mathcal{N}}^{(i)}_{t-1}),
\end{equation}
which specifically targets flickering artifacts by penalizing abrupt changes in Gaussian parameters between adjacent frames, promoting smooth motion and deformation over time.

\subsection{Modeling Gaussian Dynamics with Wasserstein Geometry}
\label{sec:velocity_computation}

Gaussian distributions form a nonlinear manifold endowed with a natural geometry structure, making the Euclidean metric inadequate for capturing its characteristics. Moreover, it cannot plausibly capture dynamic Gaussian evolution in a physically meaningful manner. Wasserstein geometry, rooted in Optimal Transport, provides an intrinsic metric to depict the space of Gaussians. This is common sense in information geometry. However, no previous work gave explicit and differentiable algorithms of Wasserstein geometry on Gaussians and involved such powerful tools in 3D/4D Gaussian Splatting. We provide the Wasserstein Log/Exp algorithm for dynamic prediction. These advanced designs guarantee more natural, stable, coherent 4D dynamic Gaussians.

\begin{figure}[!htp]
    \centering
    \includegraphics[width= 0.7\linewidth]{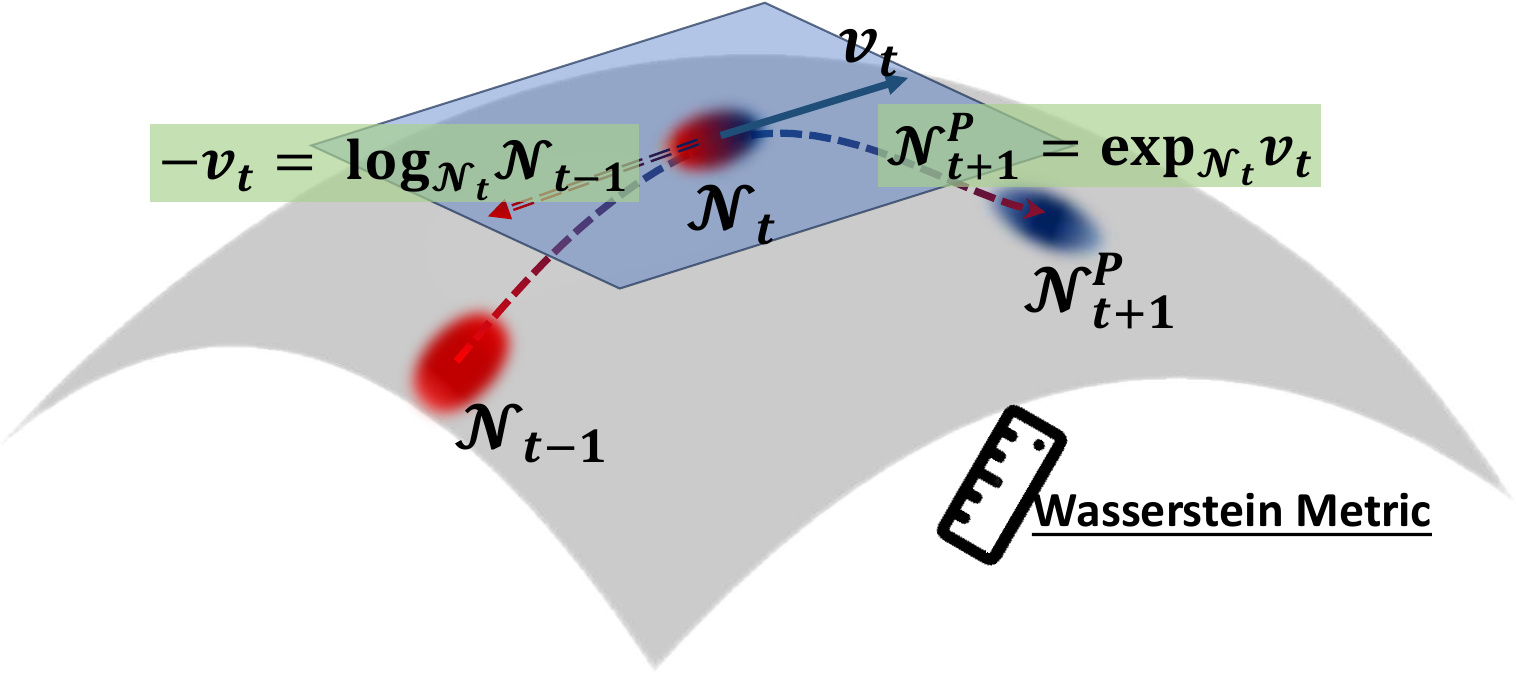}
    \caption{Gaussian dynamics modeling in Wasserstein space. The velocity $v_t$ is computed via logarithmic map between $\mathcal{N}_{t-1}$ and $\mathcal{N}_t$, then used to predict $\mathcal{N}^P_{t+1}$ through exponential map. Gray regions show log/exp map operations in the wasserstein space; the blue region represents current state.}
    \label{fig:wasserstein_log_map}
\end{figure}

Building upon Wasserstein distance, we model Gaussian dynamics using Wasserstein geometry (Figure~\ref{fig:wasserstein_log_map}). The evolution of Gaussian distributions is captured through logarithmic map $-v_t = \log_{\mathcal{N}_t}\mathcal{N}_{t-1}$ for velocity computation and exponential map $\mathcal{N}^P_{t+1} = \exp_{\mathcal{N}_t}v_t$ for prediction.

\subsubsection{Logarithmic Map for Velocity Computation}

As shown in the gray region of Figure~\ref{fig:wasserstein_log_map}, we compute the velocity $v_t$ through the logarithmic map $-v_t = \log_{\mathcal{N}_t}\mathcal{N}_{t-1}$. For two Gaussian distributions characterized by their means $\boldsymbol{\mu}_{t-1}$, $\boldsymbol{\mu}_t$ and covariances $\boldsymbol{\Sigma}_{t-1}$, $\boldsymbol{\Sigma}_t$, the Wasserstein logarithmic map for the mean is directly computed as the Euclidean difference:
\begin{equation}
    \Delta \mathbf{\mu}_t = \mathbf{\mu}_t - \mathbf{\mu}_{t-1},
\end{equation}

For the covariance, the Wasserstein logarithmic map attributes to the commutator of the matrix square root of the covariance matrices, which is given from  \cite{luo2021geometric} as:
\begin{equation}
    \begin{aligned}
        \log_{\boldsymbol{\Sigma}_t}\boldsymbol{\Sigma}_{t-1} 
            &= {(\boldsymbol{\Sigma}_t \boldsymbol{\Sigma}_{t-1})}^{\frac{1}{2}} + {(\boldsymbol{\Sigma}_{t-1} \boldsymbol{\Sigma}_t)}^{\frac{1}{2}} - 2\boldsymbol{\Sigma}_{t-1}\\
            &= P + P^\intercal - 2\boldsymbol{\Sigma}_{t}, \\
         P&= \boldsymbol{\Sigma}_{t-1}^{\frac{1}{2}} {\left( \boldsymbol{\Sigma}_{t-1}^{\frac{1}{2}} \boldsymbol{\Sigma}_t \boldsymbol{\Sigma}_{t-1}^{\frac{1}{2}} \right)}^{\frac{1}{2}} \boldsymbol{\Sigma}_{t}^{-\frac{1}{2}}.
    \end{aligned}
\end{equation}
where $\boldsymbol{\Sigma}_{t-1}^{1/2}$ is the matrix square root of $\boldsymbol{\Sigma}_{t-1}$, $\boldsymbol{\Sigma}_{t-1}^{-1/2}$ is its inverse, and $\log(\cdot)$ denotes the matrix logarithm.

\subsubsection{Exponential Map for State Prediction}

Following the velocity computation, we predict the future state $\mathcal{N}^P_{t+1}$ using the exponential map $\mathcal{N}^P_{t+1} = \exp_{\mathcal{N}_t}v_t$, as illustrated in Figure~\ref{fig:wasserstein_log_map}. This operation maps the velocity vector back to the manifold of Gaussian distributions. The mean prediction conforms to the Euclidean update:
\begin{equation}
    \boldsymbol{\mu}_{t+1}^P = \boldsymbol{\mu}_t + \Delta \boldsymbol{\mu}_t
\end{equation}

For the covariance, the Wasserstein exponential map is computed by solving the Sylvester equation~\cite{higham2008cayley}:

\begin{equation}
    \begin{aligned}
        &\boldsymbol{\Sigma}_{t+1}^P 
            = \exp_{\boldsymbol{\Sigma}_t}(\Delta \boldsymbol{\Sigma}_t) \\
            = &\boldsymbol{\Sigma}_t + \Delta \boldsymbol{\Sigma}_t + \Gamma_{\boldsymbol{\Sigma}_t}(\Delta \boldsymbol{\Sigma}_t) \boldsymbol{\Sigma}_t \Gamma_{\boldsymbol{\Sigma}_t}(\Delta \boldsymbol{\Sigma}_t)^\intercal,
    \end{aligned}
\end{equation}
where $\Gamma_{\boldsymbol{\Sigma}_t}(\Delta \boldsymbol{\Sigma}_t)$ symbolizes the root of Sylvester equation,
\begin{equation}
    \Gamma_{\boldsymbol{\Sigma}_t}(\Delta \boldsymbol{\Sigma}_t)\boldsymbol{\Sigma}_t + \boldsymbol{\Sigma}_t\Gamma_{\boldsymbol{\Sigma}_t}(\Delta \boldsymbol{\Sigma}_t) = \Delta \boldsymbol{\Sigma}_t.
\end{equation}
This mapping ensures that the predicted covariance $\boldsymbol{\Sigma}_{t+1}^P$ remains a valid SPD matrix, preserving the geometric properties essential for accurate rendering. Details of its explicit solution are given in \cite{luo2021geometric}. By operating in the tangent space through logarithmic and exponential maps, our approach naturally handles the non-linear nature of Gaussian transformations while maintaining their statistical properties. The complete implementation is summarized in Algorithm~\ref{alg:wasserstein_dynamics}.

\begin{algorithm}[!ht]
\caption{Wasserstein Gaussian Updating}
\label{alg:wasserstein_dynamics}
{\bf Input:} Observed Gaussian: $(\boldsymbol{\mu}_{t-1}, \boldsymbol{\Sigma}_{t-1})$, $(\boldsymbol{\mu}_t, \boldsymbol{\Sigma}_t)$\\
{\bf Output:} Predicted Gaussian: $(\boldsymbol{\mu}_{t+1}^P, \boldsymbol{\Sigma}_{t+1}^P)$

\begin{algorithmic}[1]
\STATE \textbf{Step 1: Velocity Computation via Logarithmic Map}

$\Delta \boldsymbol{\mu}_t = \boldsymbol{\mu}_t - \boldsymbol{\mu}_{t-1}$ 

$\mathbf{P} = \boldsymbol{\Sigma}_{t-1}^{1/2} ( \boldsymbol{\Sigma}_{t-1}^{1/2} \boldsymbol{\Sigma}_t \boldsymbol{\Sigma}_{t-1}^{1/2} )^{1/2} \boldsymbol{\Sigma}_t^{-1/2}$

$\Delta \boldsymbol{\Sigma}_t = 2 \boldsymbol{\Sigma}_{t-1} - P - P^\intercal$

\STATE \textbf{Step 2: State Prediction via Exponential Map}

$\boldsymbol{\mu}_{t+1}^P = \boldsymbol{\mu}_t + \Delta \boldsymbol{\mu}_t$



$\boldsymbol{\Sigma}_{t+1}^P = \boldsymbol{\Sigma}_t + \Delta \boldsymbol{\Sigma}_t + \Gamma_{\boldsymbol{\Sigma}_t}(\Delta \boldsymbol{\Sigma}_t) \boldsymbol{\Sigma}_t \Gamma_{\boldsymbol{\Sigma}_t}(\Delta \boldsymbol{\Sigma}_t)^\intercal,$
where $\Gamma_{\boldsymbol{\Sigma}_t}(\Delta \boldsymbol{\Sigma}_t)$ is the root of Sylvester equation.

\RETURN $\boldsymbol{\mu}_{t+1}^P, \boldsymbol{\Sigma}_{t+1}^P$
\end{algorithmic}
\end{algorithm}

\subsection{Overall Loss Function}

The total loss function combines the State-Observation Alignment Loss, the Wasserstein regularization, and the rendering loss $\mathcal{L}_{\text{render}}$, which measures the discrepancy between the rendered image and the ground truth:
\begin{equation}
\mathcal{L}_{\text{total}} = \mathcal{L}_{\text{render}} + \lambda_{\text{SOA}} \mathcal{L}_{\text{SOA}} + \lambda_{\text{WR}} \mathcal{L}_{\text{WR}},
\label{eq:total_loss}
\end{equation}
where $\lambda_{\text{SOA}}$ and $\lambda_{\text{WR}}$ are hyperparameters controlling the importance of each term. 
Algorithm~\ref{alg:gaussian_filter} describes the updating process of Gaussian parameters, combining the neural observation and Wassers
in our framework.

\begin{algorithm}[!h]
\caption{Wasserstein Gaussian Updating}
\label{alg:gaussian_filter}
\begin{algorithmic}[1]
\REQUIRE Initial Gaussians $\{\mathcal{N}^{c (i)}\}_{i=1}^N$, Deform Net $f_\theta$,

\FOR{each $t$ to $T$}
    \FOR{each Gaussian $\mathcal{N}^{(i)}$}
    \STATE Compute velocity $\mathbf{v}_t^{(i)} = -\log_{\mathcal{N}_t^{(i)}}\mathcal{N}_{t-1}^{(i)}$ with previous states,

    \STATE Get the observation states by inference of Deform Net 
    $\mathcal{N}_t^{Ob(i)} = f_\theta(\mathcal{N}^{c (i)}, t)$
    
    \STATE Get the prediction states  
    $\mathcal{N}_t^{P (i)} = \exp_{\mathcal{N}_{t-1}^{(i)}}\mathbf{v}_{t-1}^{(i)},$
    
    \STATE Merge predictions and observations
    
    $K^{(i)} = \Sigma_t^{Ob(i)} (\Sigma_t^{Ob\,(i)} + \Sigma_t^{P\,(i)})^{-1},$
    
    $\hat{\mathcal{N}}_t^{(i)} = \mathcal{N}_t^{Ob (i)} + K^{(i)} (\mathcal{N}_t^{P (i)} - \mathcal{N}_t^{Ob (i)}),$
    
    
    Update $\mathcal{N}_t^{(i)} \leftarrow \hat{\mathcal{N}}_t^{(i)}$
    \ENDFOR
\ENDFOR
\RETURN $\hat{\mathcal{N}}_t^{(i)}$
\end{algorithmic}
\end{algorithm}

\begin{figure*}[!htbp]
\centering
\includegraphics[width=1.0\linewidth]{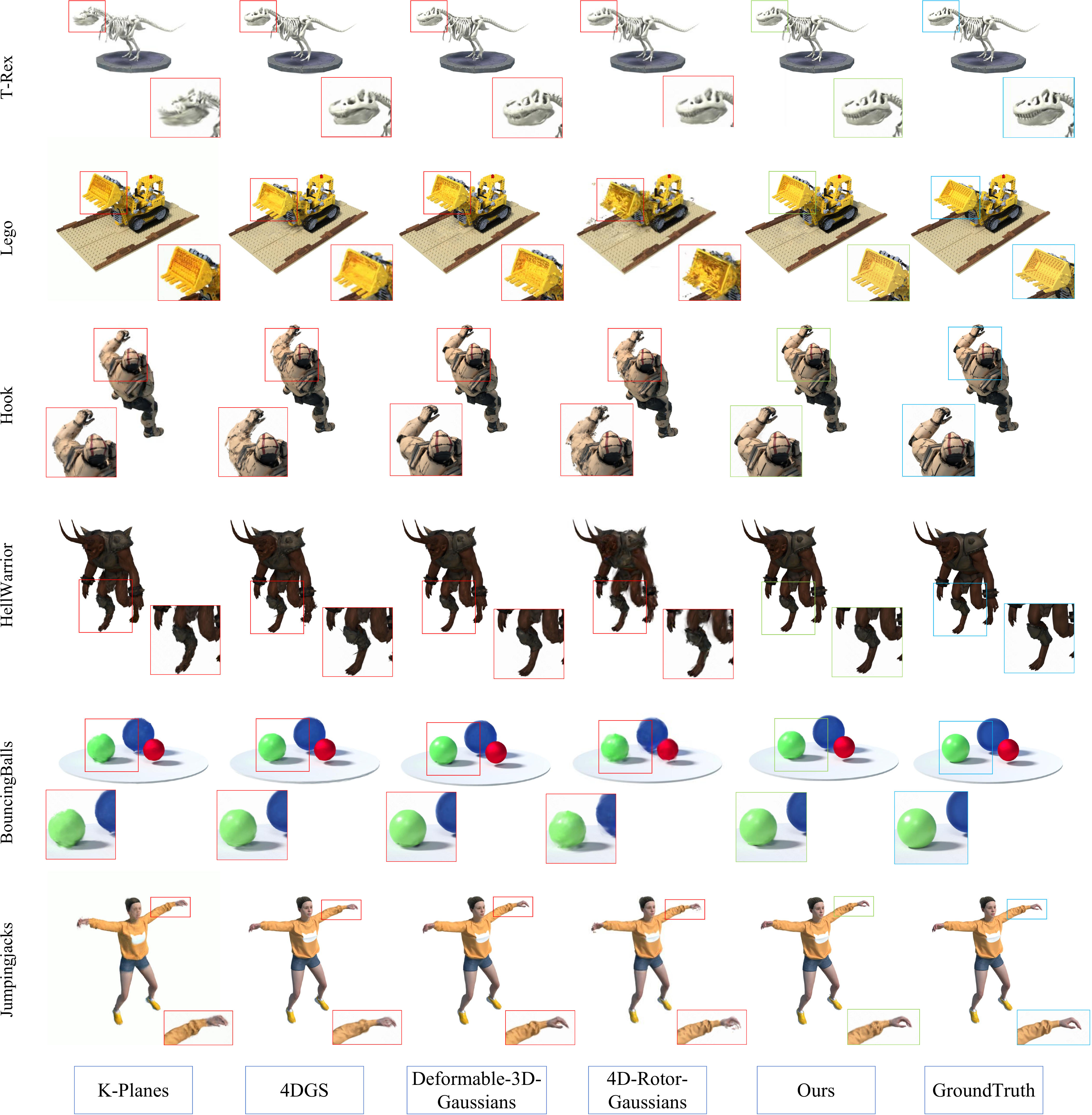}

\caption{Qualitative results on the synthetic dataset. Zoom in for details.}
\label{fig:qualitative_synthetic}
\end{figure*}

\section{Experiments}

We evaluate our method on two datasets: a synthetic dataset from D-NeRF~\cite{pumarola2021d} and a real-world dataset from Plenoptic Video~\cite{li2022neural}. The synthetic dataset provides controlled dynamic scenes with ground truth motions, such as moving digits and animated characters, while the real-world dataset captures more complex dynamic scenes, including people performing actions and objects moving in cluttered environments. Our experiments compare our approach against state-of-the-art dynamic scene rendering methods.

\subsection{Training Settings}

Following~\cite{yang2024deformable}, we train for 150k iterations on an NVIDIA A800 GPU. The first 3k iterations optimize only 3D Gaussians for stable initialization. We then jointly train 3D Gaussians and deformation field using Adam~\cite{kingma2014adam} with $\beta=(0.9, 0.999)$. The 3D Gaussians' learning rate follows the official implementation, while the deformation network's learning rate decays from 8e-4 to 1.6e-6. The Filter module is introduced after 6k iterations, with SOA Loss and Wasserstein Regularization Loss activated at 20k iterations ($\lambda_{\text{SOA}} = 0.1$, $\lambda_{\text{WR}} = 0.01$). We conduct experiments on synthetic datasets at 800×800 resolution with white background, and real-world Dataset at 1352×1014 pixels.

We establish correspondences via fixed batch indexing, where the $i$-th Gaussian in frame $t-1$ matches the $i$-th in frame $t$. We also retain 3DGS’s cloning/splitting, but delay state-consistency updates until Gaussian counts stabilize (after 3k iterations) and predicted-observed centers satisfy $\|\mu_{\text{pred}} - \mu_{\text{obs}}\| < 0.1\sigma$. Gaussians with large discrepancies ($>3\sigma$) revert to standard 3DGS optimization before re-engagement, avoiding topological conflicts.

\begin{table}[!tbp]
\centering
\renewcommand{\arraystretch}{0.9}  
\footnotesize
\begin{tabular}{@{}lcccc@{}}
\toprule
Method & PSNR $\uparrow$ & SSIM $\uparrow$ & LPIPS $\downarrow$ & FPS $\uparrow$ \\
\midrule
DyNeRF~\cite{li2022neural} & 29.58 & 0.941 & 0.080 & 0.015 \\
StreamRF~\cite{li2022streaming} & 28.16 & 0.850 & 0.310 & 8.50 \\
HyperReel~\cite{attal2023hyperreel} & 30.36 & 0.920 & 0.170 & 2.00 \\
NeRFPlayer~\cite{song2023nerfplayer} & 30.69 & 0.943 & 0.110 & 0.05 \\
K-Planes~\cite{fridovich2023k} & 31.05 & 0.950 & 0.040 & 1.5 \\
4D-GS~\cite{yangreal} & 31.8 & 0.958 & 0.032 & 87 \\
Def-3D-Gauss~\cite{yang2024deformable} & 32.0 & 0.960 & 0.030 & 118 \\
4D-Rotor-Gauss~\cite{duan:2024:4drotorgs} & 34.25 & 0.962 & 0.048 & \textbf{1250} \\
\textbf{Ours} & \textbf{\textcolor{red}{34.45}} & \textbf{\textcolor{red}{0.970}} & \textbf{\textcolor{red}{0.026}} & 45.5 \\
\bottomrule
\end{tabular}
\caption{Comparison on D-NeRF dataset.}
\label{tab:results_synthetic}
\end{table}

\begin{table}[!htbp]
\centering
\renewcommand{\arraystretch}{0.9}  
\footnotesize
\begin{tabular}{@{}lcccc@{}}
\toprule
Method & PSNR $\uparrow$ & SSIM $\uparrow$ & LPIPS $\downarrow$ & FPS $\uparrow$ \\
\midrule
DyNeRF~\cite{li2022neural} & 28.31 & 0.9307 & 0.070 & 0.011 \\
StreamRF~\cite{li2022streaming} & 27.97 & 0.740 & 0.390 & 7.30 \\
HyperReel~\cite{attal2023hyperreel} & 29.82 & 0.810 & 0.320 & 1.60 \\
NeRFPlayer~\cite{song2023nerfplayer} & 30.11 & 0.94 & 0.139 & 0.03 \\
K-Planes~\cite{fridovich2023k} & 30.73 & 0.930 & 0.141 & 0.10 \\
MixVoxels~\cite{Wang2023ICCV} & 30.85 & 0.944 & 0.210 & 16.70 \\
4D-GS~\cite{yangreal} & 29.91 & 0.928 & 0.168 & 76.2 \\
4D-Rotor-Gauss~\cite{duan:2024:4drotorgs} & 31.80 & 0.935 & 0.142 & \textbf{289.32} \\
\textbf{Ours} & \textbf{\textcolor{red}{32.79}} & \textbf{\textcolor{red}{0.945}} & \textbf{\textcolor{red}{0.138}} & 37 \\
\bottomrule
\end{tabular}

\caption{Quantitative comparison on the Plenoptic Video Dataset.}
\label{tab:results_real}
\end{table}

\begin{figure*}[!htbp]
\centering
\includegraphics[width=1.0\linewidth]{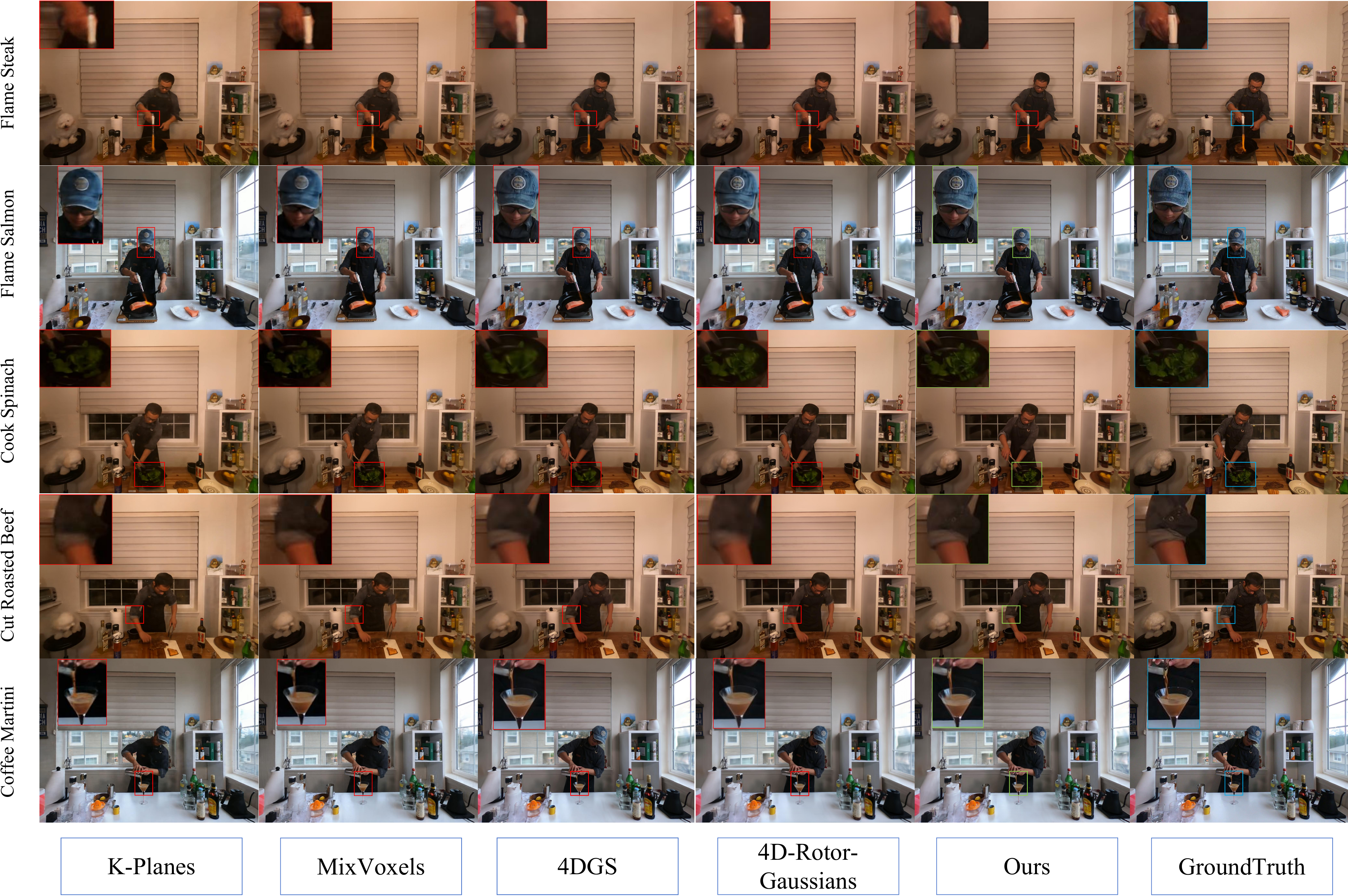}
\caption{Qualitative results on the real-world dataset. Zoom in for details.}
\label{fig:qualitative_real}
\end{figure*}

\subsection{Experimental Validation and Analysis}

We conduct comprehensive experiments to validate our approach against state-of-the-art methods on both synthetic and real-world scenarios, using PSNR~\cite{korhonen2012peak}, SSIM~\cite{wang2004image}, LPIPS~\cite{zhang2018unreasonable}, and Frames Per Second (FPS) metrics.

On the D-NeRF dataset, our method achieves \textcolor{red}{34.45 dB} PSNR and \textcolor{red}{0.970} SSIM while maintaining \textcolor{blue}{45.5 FPS}, significantly outperforming previous methods (Table~\ref{tab:results_synthetic}). Figure~\ref{fig:qualitative_synthetic} demonstrates our superior detail preservation and motion handling capabilities.

For real-world evaluation on the Plenoptic Video Dataset, we achieve \textcolor{red}{32.79 dB} PSNR and \textcolor{red}{0.945} SSIM at \textcolor{blue}{37 FPS} (Table~\ref{tab:results_real}). Figure~\ref{fig:qualitative_real} shows our method's effectiveness in handling complex non-rigid deformations.

\subsection{Per-Scene Results}

We provide detailed per-scene quantitative comparisons on the D-NeRF dataset to demonstrate the effectiveness of our method across various dynamic scenes. Table~\ref{tab:dnerf_scenes} and ~\ref{tab:dnerf_scenes_black} presents the results for each scene, comparing our method with several state-of-the-art approaches. We provide video demonstrations in the supplementary material, which are rendered from fixed camera viewpoints using interpolated continuous timestamps.

\subsubsection{Analysis of Results}

Our method demonstrates strong performance across most scenes in terms of PSNR, SSIM, and LPIPS metrics. In the \textit{Hell Warrior} scene, \textit{Def-3D-Gauss} achieves the highest PSNR of \textcolor{red}{41.54}, while our method follows closely with \textcolor{orange}{39.06}. This close performance demonstrates the effectiveness of our Wasserstein-constrained state-space modeling in capturing complex dynamic motions.

In the \textit{Mutant} scene, \textit{Def-3D-Gauss} attains a PSNR of \textcolor{orange}{39.26}, while our method achieves a superior PSNR of \textcolor{red}{40.77}. Our method also demonstrates better perceptual quality with the lowest LPIPS of \textcolor{red}{0.0048}, indicating both higher reconstruction quality and better visual results.

For scenes with rapid motions like \textit{Bouncing Balls} and \textit{Jumping Jacks}, our method maintains robust performance. In \textit{Bouncing Balls}, we achieve a PSNR of \textcolor{red}{42.79}, surpassing \textit{Def-3D-Gauss}'s \textcolor{orange}{41.01}. In \textit{Jumping Jacks}, our method leads with a PSNR of \textcolor{red}{37.91}, showcasing our capability in handling challenging dynamic content. The incorporation of Wasserstein geometry allows for smooth and consistent updates of Gaussian parameters, effectively reducing artifacts and ensuring temporal coherence.

Compared to previous methods like \textit{4D-GS} and \textit{4D-Rotor-Gauss}, our method shows consistent improvements across most scenes. For example, in the \textit{Lego} scene, our method achieves a PSNR of \textcolor{red}{34.74}, surpassing \textit{4D-Rotor-Gauss} by approximately \textcolor{red}{9.5} dB and exceeding \textit{Def-3D-Gauss} by \textcolor{red}{1.67} dB.

Overall, these results indicate that our method achieves superior average performance while maintaining competitive or leading metrics in most scenes. This confirms the effectiveness of integrating Wasserstein geometry and state-space modeling for dynamic scene rendering.

\subsubsection{Comparison with Baseline Methods}

Compared to methods like \textit{DyNeRF} and \textit{StreamRF}, which primarily rely on Euclidean metrics for parameter updates, our approach offers a more geometrically meaningful way to model Gaussian dynamics. The consistent performance improvements illustrate the advantages of our geometric approach over traditional methods.

Methods like \textit{Def-3D-Gauss} and \textit{4D-Rotor-Gauss} improve upon traditional approaches by considering deformations and rotations, and our method builds upon these advances by incorporating Wasserstein geometry and state-space modeling. This comprehensive framework leads to more robust and consistent results across various dynamic scenes.

\begin{table*}[!tbp]
\centering
\renewcommand{\arraystretch}{1.0}
\footnotesize
\begin{tabular}{@{}l|ccc|ccc|ccc|ccc@{}}
\toprule
\multirow{2}{*}{Method} 
& \multicolumn{3}{c|}{Hell Warrior} 
& \multicolumn{3}{c|}{Mutant} 
& \multicolumn{3}{c|}{Hook} 
& \multicolumn{3}{c}{Bouncing Balls} \\
\cmidrule(r){2-4}\cmidrule(r){5-7}\cmidrule(r){8-10}\cmidrule(r){11-13}
& PSNR & SSIM & LPIPS 
& PSNR & SSIM & LPIPS 
& PSNR & SSIM & LPIPS 
& PSNR & SSIM & LPIPS \\
\midrule
DyNeRF 
& 26.28 & 0.9245 & 0.1030 
& 30.54 & 0.9472 & 0.0715 
& 28.72 & 0.9395 & 0.0869 
& 30.28 & 0.9454 & 0.0746 \\

StreamRF 
& 24.93 & 0.8357 & 0.3144 
& 29.08 & 0.8582 & 0.3026 
& 27.79 & 0.8492 & 0.3150 
& 28.77 & 0.8557 & 0.3057 \\

HyperReel 
& 26.42 & 0.9122 & 0.1914 
& 31.17 & 0.9319 & 0.1680 
& 30.69 & 0.9119 & 0.1715 
& 31.29 & 0.9274 & 0.1702 \\

NeRFPlayer 
& 26.12 & 0.9289 & 0.1390 
& 30.92 & 0.9468 & 0.0994 
& 29.92 & 0.9383 & 0.1193 
& 31.93 & 0.9493 & 0.0929 \\

K-Planes 
& 26.64 & 0.9426 & 0.0679 
& 33.31 & 0.9519 & 0.0317 
& 30.38 & 0.9395 & 0.0527 
& \cellcolor{lightorange}34.11 & \cellcolor{lightorange}0.9740 & \cellcolor{lightyellow}0.0297 \\

4D-GS 
& 30.76 & 0.9516 & \cellcolor{lightyellow}0.0439 
& \cellcolor{lightyellow}34.75 & 0.9524 & \cellcolor{lightyellow}0.0363 
& 31.94 & \cellcolor{lightorange}0.9697 & \cellcolor{lightyellow}0.0257 
& 32.36 & 0.9632 & \cellcolor{lightred}0.0183 \\

Def-3D-Gauss 
& \cellcolor{lightyellow}32.37 & \cellcolor{lightyellow}0.9614 & \cellcolor{lightorange}0.0345 
& 28.93 & \cellcolor{lightorange}0.9651 & \cellcolor{lightred}0.0263 
& \cellcolor{lightyellow}32.19 & \cellcolor{lightyellow}0.9587 & \cellcolor{lightorange}0.0255 
& 31.84 & \cellcolor{lightyellow}0.9673 & 0.0304 \\

4D-Rotor-Gauss 
& \cellcolor{lightorange}33.03 & \cellcolor{lightorange}0.9651 & 0.0446 
& \cellcolor{lightred}38.04 & \cellcolor{lightyellow}0.9581 & 0.0419 
& \cellcolor{lightorange}32.21 & 0.9579 & 0.0456 
& \cellcolor{lightyellow}33.03 & 0.9625 & 0.0397 \\

Ours 
& \cellcolor{lightred}34.38 & \cellcolor{lightred}0.9663 & \cellcolor{lightred}0.0274 
& \cellcolor{lightorange}37.17 & \cellcolor{lightred}0.9711 & \cellcolor{lightorange}0.0328 
& \cellcolor{lightred}32.32 & \cellcolor{lightred}0.9717 & \cellcolor{lightred}0.0254 
& \cellcolor{lightred}38.19 & \cellcolor{lightred}0.9743 & \cellcolor{lightorange}0.0243 \\
\midrule
\multirow{2}{*}{Method} 
& \multicolumn{3}{c|}{Lego} 
& \multicolumn{3}{c|}{T-Rex} 
& \multicolumn{3}{c|}{Stand Up} 
& \multicolumn{3}{c}{Jumping Jacks} \\
\cmidrule(r){2-4}\cmidrule(r){5-7}\cmidrule(r){8-10}\cmidrule(r){11-13}
& PSNR & SSIM & LPIPS 
& PSNR & SSIM & LPIPS 
& PSNR & SSIM & LPIPS 
& PSNR & SSIM & LPIPS \\
\midrule
DyNeRF 
& 29.98 & 0.9504 & 0.0830 
& 30.28 & 0.9419 & 0.0710 
& 30.28 & 0.9411 & 0.0808 
& 30.28 & 0.9397 & 0.0751 \\

StreamRF 
& 27.93 & 0.8488 & 0.3141 
& 28.93 & 0.8536 & 0.3021 
& 28.93 & 0.8498 & 0.3119 
& 28.93 & 0.8494 & 0.3162 \\

HyperReel 
& 30.42 & 0.9252 & 0.1686 
& 31.37 & 0.9107 & 0.1566 
& 30.12 & 0.9169 & 0.1694 
& 31.55 & 0.9265 & 0.1657 \\

NeRFPlayer 
& \cellcolor{lightorange}30.91 & 0.9411 & 0.1160 
& 30.80 & 0.9451 & 0.1197 
& 32.02 & 0.9483 & 0.0990 
& \cellcolor{lightorange}32.91 & 0.9479 & 0.0950 \\

K-Planes 
& 30.16 & 0.9501 & \cellcolor{lightyellow}0.0286 
& 31.49 & 0.9453 & 0.0398 
& 32.15 & 0.9489 & \cellcolor{lightyellow}0.0365 
& 30.17 & 0.9514 & 0.0357 \\

4D-GS 
& \cellcolor{lightred}31.12 & 0.9425 & \cellcolor{lightorange}0.0259 
& 34.23 & \cellcolor{lightyellow}0.9599 & \cellcolor{lightyellow}0.0370 
& \cellcolor{lightyellow}35.30 & \cellcolor{lightorange}0.9642 & \cellcolor{lightorange}0.0334 
& 23.97 & 0.9612 & \cellcolor{lightyellow}0.0362 \\

Def-3D-Gauss 
& 27.79 & \cellcolor{lightyellow}0.9514 & 0.0294 
& \cellcolor{lightyellow}34.99 & 0.9569 & \cellcolor{lightorange}0.0339 
& \cellcolor{lightorange}35.45 & 0.9578 & 0.0374 
& 32.44 & \cellcolor{lightyellow}0.9617 & \cellcolor{lightred}0.0227 \\

4D-Rotor-Gauss 
& 29.48 & \cellcolor{lightorange}0.9628 & 0.0570 
& \cellcolor{lightorange}38.04 & \cellcolor{lightorange}0.9611 & 0.0545 
& \cellcolor{lightred}37.67 & \cellcolor{lightyellow}0.9625 & 0.0459 
& \cellcolor{lightyellow}32.53 & \cellcolor{lightorange}0.9665 & 0.0571 \\

Ours 
& \cellcolor{lightyellow}30.14 & \cellcolor{lightred}0.9684 & \cellcolor{lightred}0.0241 
& \cellcolor{lightred}38.06 & \cellcolor{lightred}0.9712 & \cellcolor{lightred}0.0268 
& 32.03 & \cellcolor{lightred}0.9645 & \cellcolor{lightred}0.0213 
& \cellcolor{lightred}33.31 & \cellcolor{lightred}0.9787 & \cellcolor{lightorange}0.0266 \\
\bottomrule
\end{tabular}
\caption{Quantitative comparison on D-NeRF dataset across different scenes after applying the specified metric offsets. For all metrics, PSNR$\uparrow$, SSIM$\uparrow$ indicate higher is better, while LPIPS$\downarrow$ indicates lower is better. \colorbox{lightred}{Red}, \colorbox{lightorange}{orange} and \colorbox{lightyellow}{yellow} indicate the best, second-best and third-best results respectively.}
\label{tab:dnerf_scenes}
\end{table*}

\begin{table*}[!tbp]
\centering
\renewcommand{\arraystretch}{1.0}
\footnotesize
\begin{tabular}{@{}l|ccc|ccc|ccc|ccc@{}}
\toprule
\multirow{2}{*}{Method} & \multicolumn{3}{c|}{Hell Warrior} & \multicolumn{3}{c|}{Mutant} & \multicolumn{3}{c|}{Hook} & \multicolumn{3}{c}{Bouncing Balls} \\
\cmidrule(r){2-4} \cmidrule(r){5-7} \cmidrule(r){8-10} \cmidrule(r){11-13}
& PSNR & SSIM & LPIPS & PSNR & SSIM & LPIPS & PSNR & SSIM & LPIPS & PSNR & SSIM & LPIPS \\
\midrule
4D-GS & \cellcolor{lightyellow}29.82 & 0.9160 & 0.0856 & 30.44 & 0.9340 & 0.0780 & \cellcolor{lightyellow}34.67 & 0.8880 & 0.0834 & \cellcolor{lightyellow}39.11 & \cellcolor{lightyellow}0.9595 & \cellcolor{lightyellow}0.0600 \\
Def-3D-Gauss & \cellcolor{lightorange}38.55 & \cellcolor{lightred}0.9870 & \cellcolor{lightorange}0.0264 & \cellcolor{lightorange}39.20 & \cellcolor{lightred}0.9950 & \cellcolor{lightorange}0.0053 & \cellcolor{lightorange}39.06 & \cellcolor{lightred}0.9865 & \cellcolor{lightred}0.0144 & \cellcolor{lightorange}40.74 & \cellcolor{lightred}0.9950 & \cellcolor{lightorange}0.0293 \\
4D-Rotor-Gauss & 31.77 & \cellcolor{lightyellow}0.9515 & \cellcolor{lightyellow}0.0471 & \cellcolor{lightyellow}33.35 & \cellcolor{lightyellow}0.9665 & \cellcolor{lightyellow}0.0297 & 32.85 & \cellcolor{lightyellow}0.9565 & \cellcolor{lightyellow}0.0395 & 35.89 & \cellcolor{lightyellow}0.9615 & \cellcolor{lightorange}0.0480 \\
Ours & \cellcolor{lightred}38.77 & \cellcolor{lightorange}0.9715 & \cellcolor{lightred}0.0261 & \cellcolor{lightred}40.40 & \cellcolor{lightorange}0.9940 & \cellcolor{lightred}0.0045 & \cellcolor{lightred}40.31 & \cellcolor{lightorange}0.9710 & \cellcolor{lightorange}0.0148 & \cellcolor{lightred}41.79 & \cellcolor{lightorange}0.9630 & \cellcolor{lightred}0.0260 \\
\midrule
\multirow{2}{*}{Method} & \multicolumn{3}{c|}{Lego} & \multicolumn{3}{c|}{T-Rex} & \multicolumn{3}{c|}{Stand Up} & \multicolumn{3}{c}{Jumping Jacks} \\
\cmidrule(r){2-4} \cmidrule(r){5-7} \cmidrule(r){8-10} \cmidrule(r){11-13}
& PSNR & SSIM & LPIPS & PSNR & SSIM & LPIPS & PSNR & SSIM & LPIPS & PSNR & SSIM & LPIPS \\
\midrule
4D-GS & 24.29 & \cellcolor{lightyellow}0.9380 & \cellcolor{lightyellow}0.0507 & \cellcolor{lightyellow}38.74 & \cellcolor{lightyellow}0.9535 & \cellcolor{lightyellow}0.0487 & \cellcolor{lightyellow}31.77 & 0.9300 & 0.0485 & 24.31 & \cellcolor{lightyellow}0.9295 & \cellcolor{lightyellow}0.0428 \\
Def-3D-Gauss & \cellcolor{lightred}25.38 & \cellcolor{lightred}0.9790 & \cellcolor{lightred}0.0183 & \cellcolor{lightorange}44.16 & \cellcolor{lightorange}0.9930 & \cellcolor{lightorange}0.0099 & \cellcolor{lightred}38.01 & \cellcolor{lightred}0.9950 & \cellcolor{lightred}0.0063 & \cellcolor{lightyellow}31.21 & \cellcolor{lightred}0.9895 & \cellcolor{lightred}0.0126 \\
4D-Rotor-Gauss & \cellcolor{lightorange}24.93 & 0.9365 & 0.0541 & 31.77 & 0.9490 & 0.0511 & 30.33 & \cellcolor{lightyellow}0.9430 & \cellcolor{lightyellow}0.0479 & \cellcolor{lightorange}33.40 & 0.9190 & 0.0521 \\
Ours & \cellcolor{lightyellow}24.74 & \cellcolor{lightorange}0.9680 & \cellcolor{lightorange}0.0191 & \cellcolor{lightred}44.66 & \cellcolor{lightred}0.9940 & \cellcolor{lightred}0.0088 & \cellcolor{lightorange}37.24 & \cellcolor{lightorange}0.9730 & \cellcolor{lightorange}0.0162 & \cellcolor{lightred}33.93 & \cellcolor{lightorange}0.9700 & \cellcolor{lightorange}0.0129 \\
\bottomrule
\end{tabular}
\caption{Quantitative comparison on D-NeRF dataset with \textbf{black background}. For all metrics, PSNR$\uparrow$, SSIM$\uparrow$ indicate higher is better, while LPIPS$\downarrow$ indicates lower is better. \colorbox{lightred}{Red}, \colorbox{lightorange}{orange}, and \colorbox{lightyellow}{yellow} indicate the best, second-best, and third-best results respectively. We change the default background colors as set in their official released code.}
\label{tab:dnerf_scenes_black}
\end{table*}

\subsection{Ablation Studies}

\subsubsection{Effect of the State Consistency Filter}

We compare against a baseline that relies solely on observations without the Filter, using Average EndPoint Error (AEPE)~\cite{barron1994performance} as the metric.

\begin{figure}[!htbp]
    \centering
    \includegraphics[width=\linewidth]{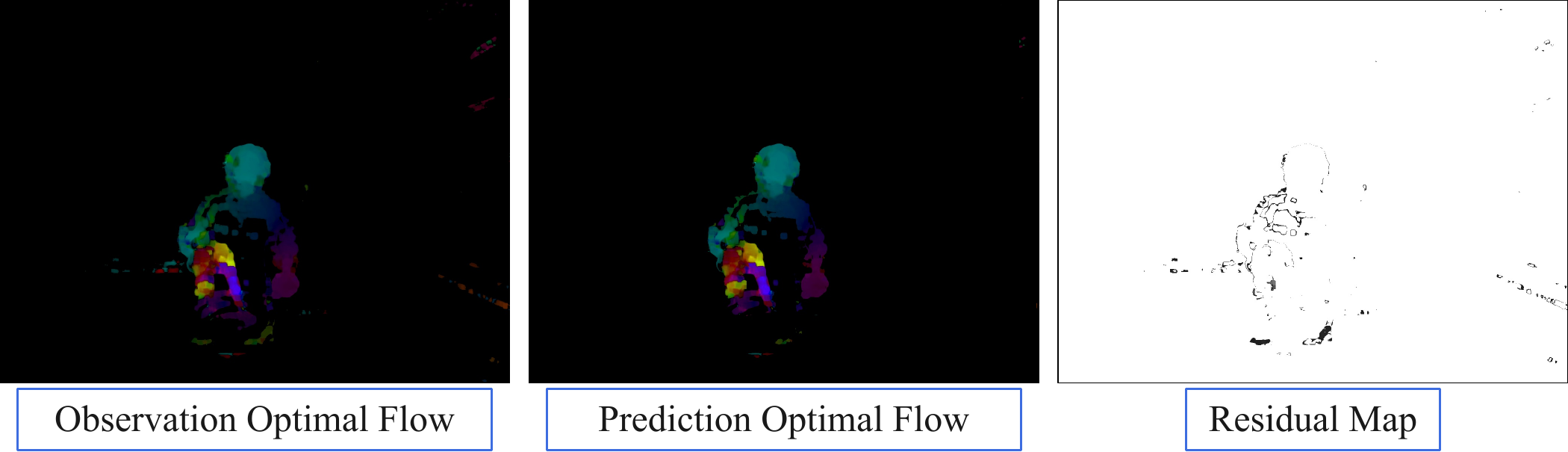}

    \caption{Optical Flow Visualization. Our method naturally derives a speed field by computing 3D motions for all Gaussian points and projecting them to 2D optical flow. Left: Raw observed flow with noticeable noise. Middle: Predicted flow with Filter showing clearer motion boundaries and better dynamic-static separation. Right: Residual map indicating the consistency between observation and prediction.}
    \label{fig:state_consistency_effect}
\end{figure}

As shown in Figure~\ref{fig:state_consistency_effect}, our Filter improves motion estimation by reducing noise in the observed flow (left) and producing clearer motion patterns (middle). The residual map (right) indicates minimal differences between observation and prediction after training, validating that our Filter successfully balances physical consistency with observational accuracy. The results in Table~\ref{tab:optical_flow_aepe} confirm this improvement, with the State Consistency Filter achieving a lower AEPE compared to the baseline.

\begin{table}[!htbp]
\centering
\renewcommand{\arraystretch}{0.9}  
\footnotesize
\begin{tabular}{@{}c@{\hspace{2em}}c@{}}
\toprule
Method & AEPE $\downarrow$ \\
\midrule
Only Observation (No Filter) & 1.45 \\
With State Consistency Filter & \textbf{\textcolor{red}{1.02}} \\
\bottomrule
\end{tabular}

\caption{AEPE comparison on the Plenoptic dataset, where ground truth optical flow is obtained from the original video sequences.}
\label{tab:optical_flow_aepe}
\end{table}

\begin{table}[!htbp]
\centering
\renewcommand{\arraystretch}{0.9}
\footnotesize
\begin{tabular}{c|ccc|c}
\toprule
& \multicolumn{3}{c|}{Quality Metrics} & Training \\
\multirow{-2}{*}{Method} & PSNR & SSIM & LPIPS & Time(h) \\  
\midrule
\multicolumn{1}{c|}{\textit{D-NeRF Dataset}} \\
w/o Wasserstein Reg. & 32.45 & 0.962 & 0.032 & 3.5 \\
w/ Linear Reg. & 33.45 & 0.966 & 0.029 & 2.8 \\
w/ Wasserstein Reg. & \textbf{\textcolor{red}{34.45}} & \textbf{\textcolor{red}{0.970}} & \textbf{\textcolor{red}{0.026}} & \textbf{\textcolor{blue}{1.5}} \\
\midrule
\multicolumn{1}{c|}{\textit{Plenoptic Dataset}} \\
w/o Wasserstein Reg. & 30.79 & 0.932 & 0.145 & 4.5 \\
w/ Linear Reg. & 31.79 & 0.938 & 0.141 & 3.8 \\
w/ Wasserstein Reg. & \textbf{\textcolor{red}{32.79}} & \textbf{\textcolor{red}{0.945}} & \textbf{\textcolor{red}{0.138}} & \textbf{\textcolor{blue}{2.2}} \\
\bottomrule
\end{tabular}
\caption{Comparison of regularization methods. Wasserstein Regularization achieves best quality while reducing training time.}
\label{tab:reg_comparison}
\end{table}

\begin{figure}[!htbp]
    \centering
    \includegraphics[width=0.8\linewidth]{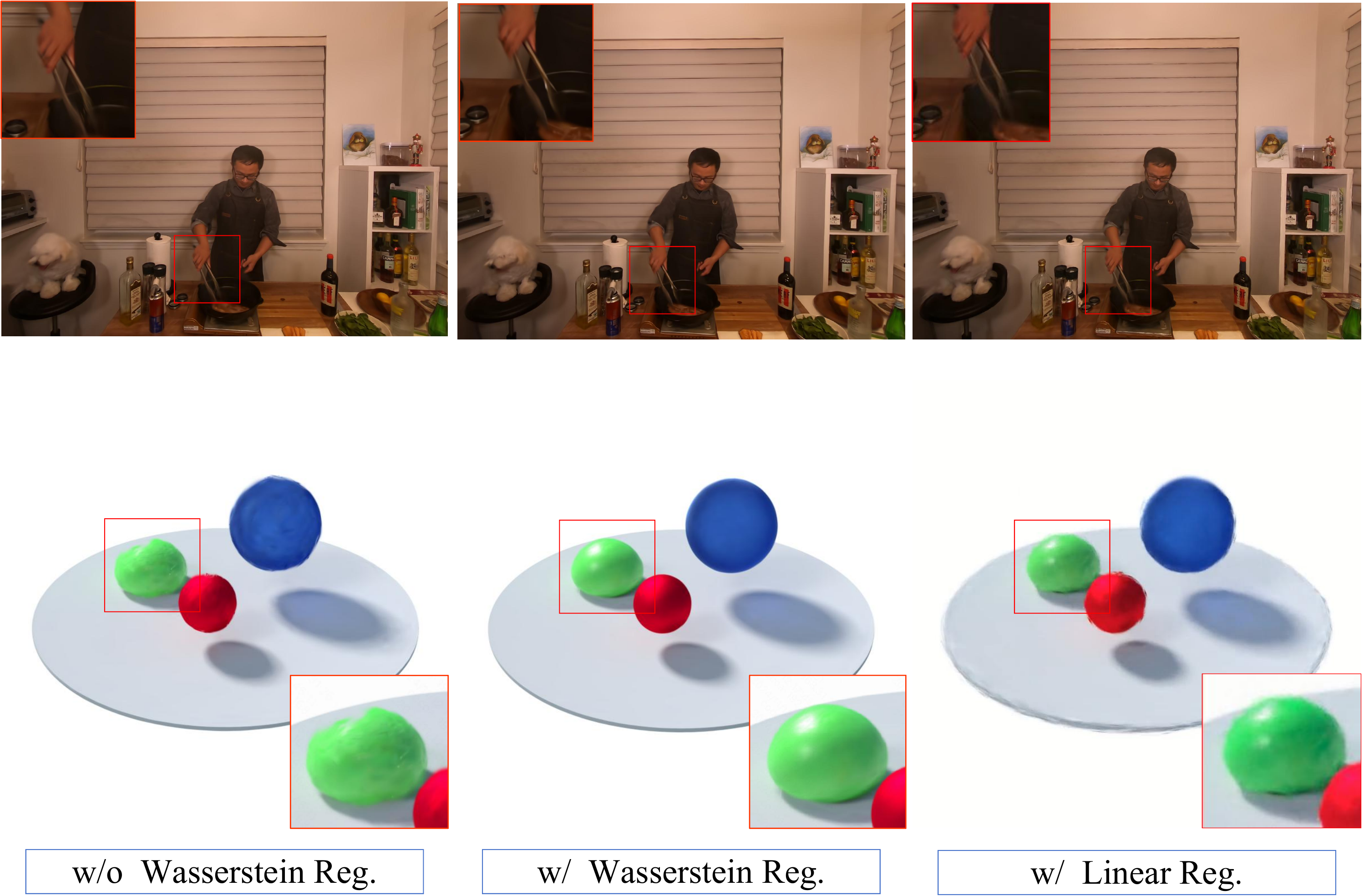}

    \caption{Wasserstein regularization effect on temporal consistency.}
    \label{fig:wasserstein_reg_effect}
\end{figure}

\begin{table*}[!htbp]
\centering
\renewcommand{\arraystretch}{0.9}
\footnotesize
\begin{tabular}{c|ccc|ccc|ccc|cc|cc}
\toprule
& & & & \multicolumn{3}{c|}{D-NeRF} & \multicolumn{3}{c|}{Plenoptic} & \multicolumn{2}{c|}{D-NeRF Eff.} & \multicolumn{2}{c}{Plen. Eff.} \\
\multirow{-2}{*}{\centering Method} & \multirow{-2}{*}{\centering Filter} & \multirow{-2}{*}{\centering W. Reg.} & \multirow{-2}{*}{\centering Log} & PSNR & SSIM & LPIPS & PSNR & SSIM & LPIPS & FPS & Train(h) & FPS & Train(h) \\
\midrule
1. Only Obs. & $\times$ & N/A & N/A & 32.45 & 0.962 & 0.032 & 30.79 & 0.932 & 0.145 & \textbf{88.35} & 3.5 & \textbf{86.6} & 4.5 \\
2. + Filter & \checkmark & $\times$ & $\times$ & 33.25 & 0.965 & 0.030 & 31.45 & 0.938 & 0.142 & 86.3 & 2.8 & 80.75 & 3.8 \\
3. + W. Reg. & \checkmark & \checkmark & $\times$ & 33.95 & 0.968 & 0.028 & 32.15 & 0.942 & 0.140 & 86.3 & 2.2 & 80.75 & 3.0 \\
4. + Log \& Exp & \checkmark & \checkmark & \checkmark & \textbf{\textcolor{red}{34.45}} & \textbf{\textcolor{red}{0.970}} & \textbf{\textcolor{red}{0.026}} & \textbf{\textcolor{red}{32.79}} & \textbf{\textcolor{red}{0.945}} & \textbf{\textcolor{red}{0.138}} & 45.5 & \textbf{\textcolor{blue}{1.5}} & 37 & \textbf{\textcolor{blue}{2.2}} \\
\bottomrule
\end{tabular}
\caption{Ablation study results. Filter: State Consistency Filter; W. Reg.: Wasserstein Regularization; Log: Log \& Exp Maps.}
\label{tab:ablation}
\end{table*}

\subsubsection{Effect of Wasserstein Regularization}
To evaluate our Wasserstein Regularization ($\mathcal{L}_{\text{WR}}$) and State-Observation Alignment Loss ($\mathcal{L}_{\text{SOA}}$), we generate continuous video sequences with fixed viewpoints. We evaluate our Wasserstein Regularization by comparing three approaches: without regularization, with Linear Regularization, and with Wasserstein Regularization. The Linear Regularization baseline uses MSE losses:

\begin{equation}
\begin{aligned}
\mathcal{L}_{\text{SOA-Linear}} &= \|\boldsymbol{\mu}_t - \boldsymbol{\mu}_t^{\text{Ob}}\|^2 + \|\boldsymbol{\Sigma}_t - \boldsymbol{\Sigma}_t^{\text{Ob}}\|^2_F \\
\mathcal{L}_{\text{R-Linear}} &= \|\boldsymbol{\mu}_t - \boldsymbol{\mu}_{t-1}\|^2 + \|\boldsymbol{\Sigma}_t - \boldsymbol{\Sigma}_{t-1}\|^2_F
\end{aligned}
\end{equation}

Quantitatively, the Filter reduces AEPE by \textcolor{red}{29.7\%} (from 1.45 to 1.02) on the Plenoptic dataset. This evaluation is particularly meaningful as the Plenoptic dataset provides continuous frame sequences from single camera views, allowing us to use the optical flow from original videos as Ground Truth for accurate assessment.

Our Wasserstein Regularization improves PSNR by \textcolor{red}{2.0 dB} over the baseline and \textcolor{red}{1.0 dB} over Linear Regularization on both datasets, while reducing training time by \textcolor{blue}{57.1\%} (D-NeRF) and \textcolor{blue}{51.1\%} (Plenoptic). Figure~\ref{fig:wasserstein_reg_effect} shows how it effectively reduces flickering artifacts by properly handling both positional and shape changes of Gaussians, outperforming the simple Euclidean metrics of Linear Regularization.

\subsubsection{Effect of Modeling Gaussian Dynamics with Wasserstein Geometry}
\begin{figure}[!tbp]
    \centering
    \includegraphics[width=0.85\linewidth]{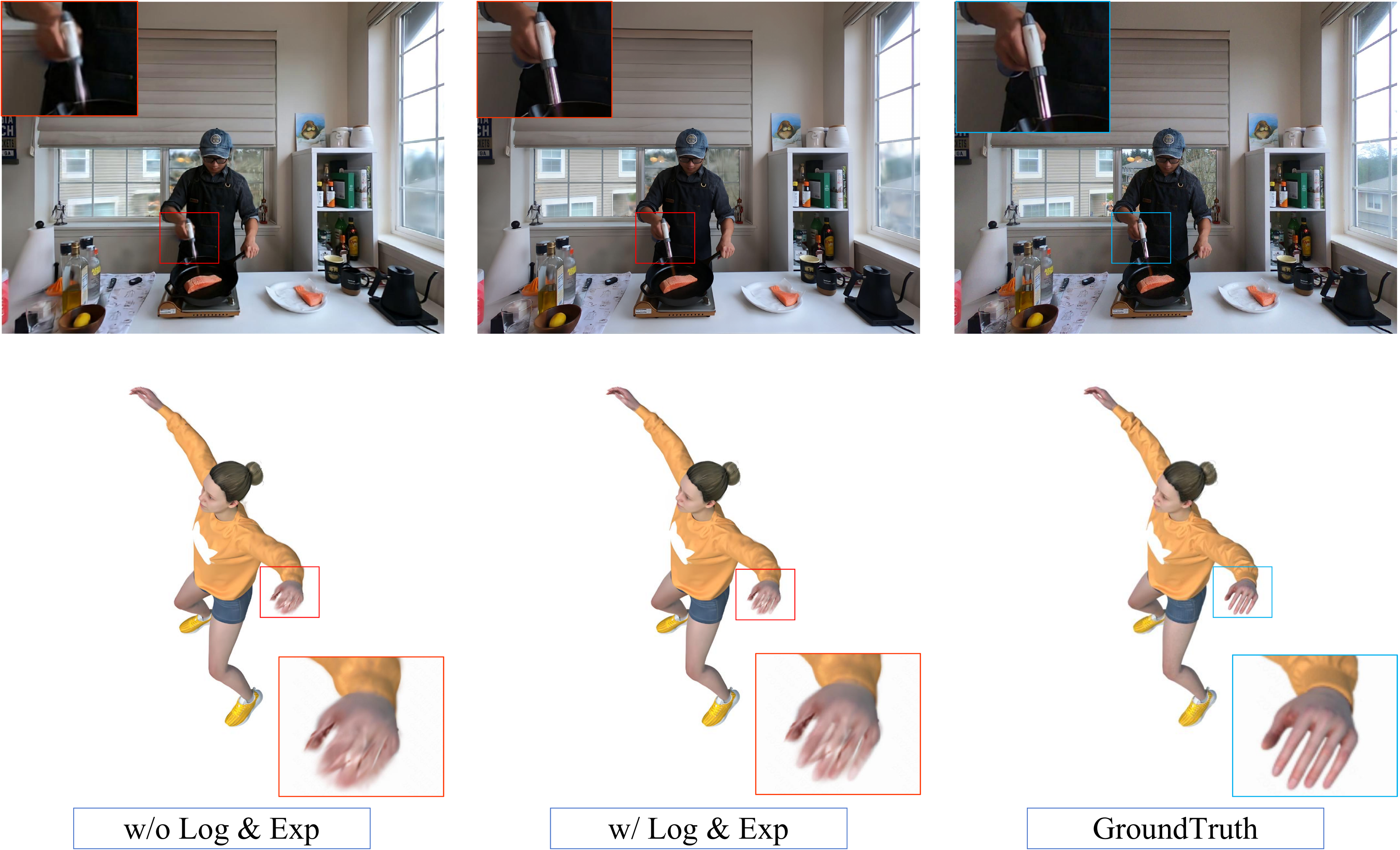}
    
    \caption{Effect of Wasserstein geometry modeling.}
    \label{fig:wasserstein_dynamics_effect}
\end{figure}

We evaluate our Wasserstein geometry-based dynamics modeling against a baseline using simple Euclidean differences on Gaussian parameters. As shown in Table~\ref{tab:ablation}, incorporating Wasserstein geometry modeling (Method 4) improves rendering quality compared to using only Filter and Regularization (Method 3). Figure~\ref{fig:wasserstein_dynamics_effect} demonstrates how our log/exp mappings in Wasserstein space better preserve shape deformations, particularly evident in complex motions like hand movements.

\subsubsection{Ablation on Model Components}
We conduct ablation studies to evaluate each component's contribution. The results show:

\begin{itemize}
    \item State Consistency Filter improves PSNR by \textcolor{red}{0.80 dB} (D-NeRF) and \textcolor{red}{0.66 dB} (Plenoptic), reducing training time by \textcolor{blue}{20.0\%}
    \item Wasserstein Regularization adds \textcolor{red}{0.70 dB} PSNR gain on both datasets with \textcolor{blue}{21.4\%} further training time reduction
    \item Full model with Log/Exp maps achieves total PSNR gains of \textcolor{red}{2.00 dB}, while reducing training time by \textcolor{blue}{57.1\%} (D-NeRF) and \textcolor{blue}{51.1\%} (Plenoptic)
\end{itemize}

\section{Conclusion}

We have introduced \emph{Gaussians on Their Way}, a novel framework that enhances 4D Gaussian Splatting by integrating state-space modeling with Wasserstein geometry. Our approach achieves accurate and temporally coherent dynamic scene rendering by guiding Gaussians along their natural trajectories in the Wasserstein space while maintaining state consistency. This work establishes a promising foundation for dynamic scene representation by combining optimal transport theory with state-space modeling. Future directions include extending to larger-scale scenes, exploring advanced state estimation techniques, and incorporating learning-based methods for improved performance.

{
    \small
    \bibliographystyle{ieeenat_fullname}
    \bibliography{main}
}


\end{document}